\documentclass[wcp]{jmlr}

\usepackage{longtable}

\usepackage{booktabs}
\usepackage{amsfonts}
\usepackage{bm}
\usepackage{color}
\usepackage{booktabs}
\usepackage{threeparttable}
\usepackage{natbib}
\usepackage{lineno}
\usepackage{enumerate}
\usepackage{multirow}
\usepackage{mathrsfs}
\usepackage{bbm}
\usepackage{CJK}
\usepackage{verbatim}
\usepackage{appendix}

\makeatletter
\let\Ginclude@graphics\@org@Ginclude@graphics 
\makeatother

\title[STCMTL]{Semisoft Task Clustering for Multi-Task Learning}

\author{\Name{Yuzhao Zhang} \Email{2016201698@ruc.edu.cn}\and
\Name{Yifan Sun} \Email{sunyifan@ruc.edu.cn}\\
 \addr Center for Applied Statistics, Renmin University of China, Beijing, China}

\begin{document}
	
	\maketitle
	
	\begin{abstract}
		Multi-task learning (MTL) aims to improve the performance of multiple related prediction tasks by leveraging useful information from them. Due to their flexibility and ability to reduce unknown coefficients substantially, the task-clustering-based MTL approaches have attracted considerable attention. Motivated by the idea of semisoft clustering of data, we propose a semisoft task clustering approach, which can simultaneously reveal the task cluster structure for both pure and mixed tasks as well as select the relevant features. The main assumption behind our approach is that each cluster has some pure tasks, and each mixed task can be represented by a linear combination of pure tasks in different clusters. To solve the resulting non-convex constrained optimization problem, we design an efficient three-step algorithm. The experimental results based on synthetic and real-world datasets validate the effectiveness and efficiency of the proposed approach. Finally, we extend the proposed approach to a robust task clustering problem.  
	\end{abstract}
	
\section{Introduction}

The learning of multiple related tasks is often observed in real-world applications, such as computer vision [\cite{CV}], web research [\cite{web}], bioinformatics [\cite{bioinformatics}], and others. Multi-task learning (MTL) is a common machine learning method that is used to solve this problem. To be more specific, MTL learns multiple related tasks simultaneously while exploiting commonalities and differences across the tasks. This can lead to better generalization performance compared to that from learning each task separately. The key challenge in MTL is how to screen the common information across related tasks while preventing information from being shared among unrelated tasks. To tackle this challenge, a multitude of MTL approaches have been proposed in the last decade. Among these approaches, feature selection approaches [\cite{feature_selection1,feature_selection4}] and task clustering approaches [\cite{Gauss2,trace1}] have been the most commonly investigated. 

The feature selection approaches aim to select a subset of the original features to serve as the shared features for different tasks, which can be achieved via the regularization [\cite{feature_selection1}] or sparse prior [\cite{SparsePrior2}] of the coefficient matrix. Compared with other methods, the feature selection methods provide better interpretability. However, 
the assumption that all tasks share identical or similar relevant features is too strong to be valid in many applications. Moreover, information can only be shared in the original feature space, which limits the ability to share information across tasks. 

Inspired by the idea of data clustering, the task clustering approaches group tasks into several clusters with similar coefficients. 
Earlier studies focused on identifying the disjoint task clusters, which can be realized via a two-stage strategy [\cite{TwoStage2}], a Gaussian mixture model [\cite{Gauss2}], or regularizations [\cite{cluster2,representative}]. Such hard clustering may not always reflect the true clustering structure and may result in the inaccurate extraction of information shared among tasks. To this end, recent studies have focused on improvements through allowing partial overlaps between different clusters, such that each task can belong to multiple clusters [\cite{overlap2,Go-MTL,VSTG}]. This is achieved by decomposing the original coefficient matrix into a product of a matrix consisting of coefficient vectors in different clusters and a matrix containing combination weights. 

Due to their flexibility and ability to reduce unknown coefficients substantially, task-clustering approaches have attracted widespread attention in the field of MTL. Although great progress has been made in the task-clustering approaches, the existing approaches still have several limitations. First, few task clustering approaches conduct feature selection, which may affect interpretability and generalizability to some degree. Second, most task-clustering approaches assume that task parameters within a cluster lie in a low dimensional subspace, which we have little information about. As a result, two major issues would arise: the cluster structure could become non-identifiable in a latent space, and the characteristics of individual clusters would be blurred. This would limit their applications in areas such as biomedicine and economics, where the reliability and interpretability of results are especially valued. 

Motivated by the idea of the semisoft clustering of data [\cite{Zhu2019Semisoft,LOVE}], we propose a novel semisoft task clustering MTL (STCMTL) approach, which can simultaneously reveal the task cluster structure for both pure tasks and mixed tasks and select relevant features. A pure task is a task that belongs to a single cluster, while a mixed task is one that belongs to two or more clusters. The key insights of the STCMTL are as follows: (a) pure and mixed tasks co-exist (which is why the term "semisoft" is used); and (b) each mixed task can be represented by a linear combination of pure tasks in different clusters, where the linear combination weights are nonnegative numbers that sum to one (called memberships) and represent the proportions of the mixed task in the corresponding clusters. The idea of representing a task as a linear combination of a set of pure (or representative) tasks has been widely adopted in the literature [\cite{representative,robust}]. However, these approaches cannot remove the irrelevant features. VSTG-MTL [\cite{VSTG}] is the most relevant method to ours, which can simultaneously perform feature selection and learn an overlapping task cluster structure by encouraging sparsity within the column of the weight matrix. However, this method fails to classify the pure and mixed tasks. 

We formulate the semisoft task clustering problem as a constrained optimization problem. A novel and efficient algorithm is developed to solve this complex problem. Specifically, we decompose the original non-convex optimization problem into three subproblems. We then employ the semisoft clustering with pure cells (SOUP) algorithm [\cite{Zhu2019Semisoft}] that was recently proposed for data semisoft clustering  to identify the set of pure tasks and estimate the soft memberships for mixed tasks. Extensive simulation studies have demonstrated the superiority of STCMTL to natural competitors in identifying cluster structure and relevant features, estimating coefficients, and especially, time cost. The effectiveness and efficiency of  STCMTL are again validated in the four real-world data sets. In addition, STCMTL can be easily extended to other task clustering problems (e.g., robust task clustering) by replacing the SOUP algorithm with other corresponding algorithms. 

In summary, the proposed STCMTL approach has the following advantages: 
\begin{itemize}
	\item It can simultaneously reveal the task cluster structure for both pure and mixed tasks as well as perform feature selection. 
	\item It provides an identifiable and interpretable task cluster structure. 
	\item  Compared with its direct competitors, it requires much less calculation time and results in more precise feature selection, and it can be easily 
	extended to other task clustering problems. 
\end{itemize}

The rest of the paper is organized as follows. We first introduce the notations and problem formulation in Section 2. The optimization procedure is presented in Section 3. In Section 4, we campare our method's experimental results with 5 other baselines on synthetic and real-world dataset. We rewrite our algotithm to a general framework and provide an alternatives under the framework in Section 5.

\section{Formulation}
Suppose we have $T$ tasks and $D$ features. In task $i (=1,2\ldots,T)$, there are $n_i$ observations. Denote $\mathbf{X}_i=[(\mathbf{x}_i^1)^\top,\ldots, (\mathbf{x}_i^{n_i})^\top]^\top\in\mathcal{R}^{n_i\times D}$ as the input matrix with $\mathbf{x}_i^j\in \mathcal{R}^{1\times D}$ and  $\mathbf{y}_i=[y_i^1,\ldots,y_i^{n_i}]^\top\in\mathcal{R}^ {n_i\times 1}$ as the output vector, where $y_i^j \in \mathcal{R}$ for regression problems and $y_i^j \in \{-1,1\}$ for binary classification problems. For each task, consider the linear relationship between the input matrix and output vector: 
$$\mathbf{y}_i = f(\mathbf{X}_i\mathbf{w}_i),\ \ i=1,2,\ldots,T, $$
where $f$ is an identify function for regression problems and a logit function for classification problems, and $\mathbf{w}_i\in  \mathcal{R}^{D\times 1}$ is the coefficient vector for the $i$th task. Denote $\mathbf{W}=[\mathbf{w}_1, \ldots, \mathbf{w}_T]\in \mathcal{R}^{D\times T}$ as the coefficient matrix to be estimated. 

We consider a setting in which $T$ tasks form $K$($K\ll \min\{D,T\}$) clusters, and each cluster $k$ has the unique coefficient vector $\mathbf{u}_k\in \mathcal{R}^{D\times 1}$. Different from many task clustering methods that partition tasks into disjoint clusters, we allow for an overlapping cluster structure in the sense that it allows a task to be a member of multiple clusters. We call the tasks that belong to a single cluster \emph{pure tasks} and the tasks that belong to two or more clusters \emph{mixed tasks}.  We assume that the coefficient vector of each task $\mathbf{w}_i$ can be represented by the linear combination of the coefficient vector of the cluster to which this task belongs; that is, $\mathbf{w}_i=\sum_{k=1}^K v_{ki}\mathbf{u}_k$, where $v_{ki}$ is the membership weight that represents the proportions of the $i$th task belonging to cluster $k$ with $0\leq v_{ki}\leq 1$ and $\sum_{k=1}^K v_{ki}=1$. To obtain a more intuitive, and more importantly, identifiable, cluster structure, we assume that each cluster has some pure tasks. Obviously, a pure task in cluster $k$ has $v_{ki}=1$ and zeros elsewhere. Denote a full rank matrix $\mathbf{U}=[\mathbf{u}_1,\ldots,\mathbf{u}_K]\in\mathcal{R}^{D\times K}$ as the cluster coefficients and $\mathbf{V}=[\mathbf{v}_1,\ldots,\mathbf{v}_T]\in\mathcal{R}^{K\times T}$ as the membership matrix with $i$th column $\mathbf{v}_i=[v_{1i},\ldots,v_{Ki}]^\top\in\mathcal{R}^{K\times 1}$ containing the membership weight of task $i$. The above assumption enables us to write the original coefficient matrix as $\mathbf{W}=\mathbf{U}\mathbf{V}$. Formally, we formulate our approach as a constrained optimization problem: 
\begin{equation}
	\begin{array}{c}
		\displaystyle \min _{\mathbf{U}, \mathbf{V}} \sum_{i=1}^{T} \displaystyle\frac{1}{n_i} L\left(\mathbf{y}_i, \mathbf{X}_i \mathbf{U} \mathbf{v}_{i}\right) \\ 
		\text { s.t }  \prod_{i=1}^T(v_{ki}-1)=0, \text{for}\ k=1,\ldots, K, \\
		\mathbf{V}\geq 0, \mathbf{V}^\top\mathbb{I}_K=\mathbb{I}_T, \\
		\parallel \mathbf{U}\parallel_1 \leq \alpha. 
		\label{main}
	\end{array}
\end{equation}
where $L(\cdot,\cdot)$ is the empirical loss function, which is a squared loss for a regression problem and a logistic loss for a binary classification problem;
the first constraint ensures the existence of pure tasks in each cluster $k$; $\mathbf{V} \geq 0$ means that each element in matrix $\mathbf{V}$ is nonnegative; $\mathbb{I}_m$ represents a $m\times 1$ vector with all elements being 1; and $\parallel \mathbf{U}\parallel_1=\sum_{k=1}^K \parallel \mathbf{u}_k\parallel_1$ is the $l_1$ norm, which encourages the sparsity in the cluster coefficients, and $\alpha$ is the constraint parameter.

\section{Optimization and Algorithm}
Traditional constrained optimization algorithms, such as the interior point method and alternating direction method of multipliers (ADMM) method, are time consuming, and will inevitably involve tedious tuning parameters. Recently,  [\cite{Zhu2019Semisoft}] proposed SOUP, a novel semisoft clustering algorithm, which can simultaneously identify the pure and mixed samples. Extensive simulation and real-data analysis demonstrate its advantages over direct competitors. Motivated by the idea of SOUP, we develop a new optimization algorithm to solve the problem in Eq. (\ref{main}). 
As the standard alternating optimization algorithm, the new algorithm updates the membership matrix $\mathbf{V}$ and cluster coefficient matrix $\mathbf{U}$ alternately. The biggest difference is that the coefficient matrix $\mathbf{W}$ is also updated after updating $\mathbf{V}$ and $\mathbf{U}$. 
The key ingredient of the algorithm is to adopt SOUP to extract the overlapping cluster structure from the coefficient matrix $\mathbf{W}$ by identifying the set of pure tasks and then estimating the membership matrix $\mathbf{V}$. Hence, $\mathbf{W}$ is required. 
In addition, to increase the numerical stability and accelerate convergence, we modify each column of $\mathbf{W}$ separately before inputting it into SOUP (see more details in the follows). In sum, the new algorithm involves a three-step process: updating $\mathbf{V}$, $\mathbf{U}$, and $\mathbf{W}$.   

\subsection{Updating $\mathbf{V}$}
We directly apply the SOUP algorithm to the coefficient matrix $\mathbf{W}$ to reveal the cluster structure among tasks. The SOUP algorithm involves two steps: identifying the set of pure tasks and estimating the membership matrix $\mathbf{V}$. In the following, we introduce the main ideas of the two steps and provide more details in the Appendix. 

\noindent\bf{Identify the set of pure tasks}\rm. Pure tasks provide valuable information from which to recover their memberships, further guiding the estimation of the membership weights for the mixed tasks. Define a task similarity matrix: $\mathbf{S}=\mathbf{W}^\top \mathbf{W}\in\mathcal{R}^{T\times T}$. To find the set of pure tasks, SOUP exploits the special block structure formed by the pure tasks in the similarity matrix $\mathbf{S}$ to calculate a purity score for each task. After sorting the purity scores in descending order, the top $\theta$ percent of tasks are declared as pure tasks, denoted as $\mathcal{P}$, and then partitioned into $K$ clusters by $K$-means algorithm. 

\noindent\bf{Estimate the membership matrix $\mathbf{V}$}\rm. It is noticed that there is a matrix $\mathbf{A}\in\mathcal{R}^{K\times K}$, such that $\mathbf{V}=\mathbf{A}\mathbf{\Theta}$, where $\Theta\in\mathcal{R}^{K\times T}$ is the matrix consisting of the top $K$ eigenvectors of matrix $\mathbf{S}$. Because we have identified the set of pure tasks $\mathcal{P}$ and their memberships $\mathbf{V}_{\mathcal{P}}$, the desired matrix $\mathbf{A}$ can be automatically determined from $\mathbf{V}_{\mathcal{P}}=\mathbf{A}\mathbf{\Theta}_{\mathcal{P}}$, where $\mathbf{\Theta}_{\mathcal{P}}$ refers to the sub-matrix of matrix $\mathbf{\Theta}$ formed by the columns in set $\mathcal{P}$. As a result, the full membership matrix can be recovered: $\mathbf{V}=\mathbf{A}\mathbf{\Theta}$. 
\label{section:update_v}

\subsection{Updating $\mathbf{U}$}
For a fixed membership matrix $\mathbf{V}$, the optimization problem in Eq. (\ref{main}) with respect to the cluster coefficient matrix $\mathbf{U}$, becomes as follows: 
\begin{eqnarray*}
	&\displaystyle\min _{\mathbf{U}}\displaystyle \sum_{i=1}^{T} \displaystyle\frac{1}{n_i} L\left(\mathbf{y}_i, \mathbf{X}_i \mathbf{U} \mathbf{v}_{i}\right) 
	\\&\text { s.t } \parallel \mathbf{U}\parallel_1 \leq \alpha. 
\end{eqnarray*}
We transform the above constraint problem to the following regularized objective function 
\begin{equation}
	\label{U_update}
	\sum_{i=1}^{T} \frac{1}{n_{i}} L\left(\mathbf{y}_i, \mathbf{X}_i \mathbf{U} \mathbf{v}_{i}\right)+\gamma\parallel \mathbf{U}\parallel_{1},
\end{equation}
where $\gamma$ is the regularization parameter. We optimize the objective function with respect to one column vector $\mathbf{u}_k$ at a time and iteratively cycle through all the columns until the Euclidian distances between $\mathbf{u}_k$'s in two adjacent steps converge. Specifically, for $k\in\{1,\ldots,K\}$, we minimize $M(\mathbf{u}_k)$ with respect to $\mathbf{u}_k$, where 
\begin{equation}
	\label{U_reg}
	M(\mathbf{u}_k)=\sum_{i=1}^{T} \frac{1}{n_{i}} \parallel\mathbf{y}_i-\mathbf{X}_i\sum_{l\neq k}v_{li}\mathbf{u}_l-v_{ki}\mathbf{X}_i\mathbf{u}_{k}\parallel_2^2+\gamma\parallel \mathbf{u}_k\parallel_{1},
\end{equation} 
for linear regression problems with a squared loss, and 
\begin{equation}
	\begin{aligned}
		&M(\mathbf{u}_k)= \sum_{i=1}^T \frac{1}{n_i} \sum_{j=1}^{n_i}\Bigg\{\log\Bigg[\exp\left(y_i^j(\mathbf{x}_i^j)^\top \sum_{l \neq k}v_{li}\mathbf{u}_l\right)\\&+\exp\left(-y_i^j v_{ki} (\mathbf{x}_i^j)^\top \mathbf{u}_k\right)\Bigg]+ \log\left[\exp\left(-y_i^j(\mathbf{x}_i^j)^\top\sum_{l \neq k}v_{li}\mathbf{u}_l\right)\right]\Bigg\}\\&+ \gamma\parallel\mathbf{u}_k\parallel_1 ,
		\label{U_logit}
	\end{aligned}
\end{equation}
for binary classification problems with a logistic loss.  
The above problems are solved by using the coordinate descent [\cite{glmnet,logist}], a well-developed method for tackling a regularized regression model. 
\label{section:update_u}

\subsection{Updating $\mathbf{W}$}
After updating $\mathbf{V}$ and $\mathbf{U}$, we obtain a new coefficient matrix $\tilde{\mathbf{W}}=\mathbf{U}\mathbf{V}$. To increase the numerical stability and achieve faster convergence, we update each column of the coefficient matrix $\mathbf{W}$ separately with $\tilde{\mathbf{W}}$ as initial values. Specifically, the $i$th column vector $\mathbf{w}_i$ can be updated by solving the following optimization problem, which 
optimizes the following objective function  
\begin{equation}
	\min_{\mathbf{w}_i}\frac{1}{n_i} L(\mathbf{y}_i,\mathbf{X}_i \mathbf{w}_i) + \lambda_i \parallel\mathbf{w}_i\parallel_1, \label{single}
\end{equation}  
where $\lambda_i$ is the regularization parameter, which is fixed during iteration (see more details in the following). We again leverage the coordinate descent algorithm to solve the problem efficiently. Note that, in practice, we apply early stopping to the coordinate descent to update each $\mathbf{w}_i$ with a given number $(2\sim 5)$ of iterations until completion.  

\subsection{Algorithm}   
\label{alg}
Algorithm 1 summarizes the whole procedure to solve the optimization problem in Eq. (\ref{main}). To start the algorithm, it is important to find a reasonable initial value of $\mathbf{W}$. For this purpose, in each task, we learn a regularized regression or logistic regression coefficient: 
\begin{equation}
	\label{initial}
	\mathbf{w}_i^0=\min_{\mathbf{w}_i}\frac{1}{n_i} L(\mathbf{y}_i,\mathbf{X}_i \mathbf{w}_i) + \lambda_i \parallel \mathbf{w}_i\parallel_1,
\end{equation}
which can be solved with the coordinate descent method. The initial value of  $\mathbf{W}$ is given by $\mathbf{W}^0=[\mathbf{w}_1^0,\ldots, \mathbf{w}_T^0]$.  

\noindent\bf{Tuning parameters}\rm. SOUP has two tuning parameters: $\epsilon$, the fraction of most neighbors to be examined for each task; and $\theta$, the fraction of tasks that are declared as pure tasks. In practice, SOUP is robust with respect to these two parameters. Following the setup in [\cite{Zhu2019Semisoft}], we set $\theta=0.5$ and $\epsilon=0.1$ throughout this study. As for the tuning parameters $\lambda_i$ ($i=1,\ldots, T$) in Eqs. (\ref{single}-\ref{initial}) and $\gamma$ in Eq. (\ref{U_update}), cross-validation over a grid search is the commonly adopted method to select the optimal combination of parameters, but this becomes increasingly prohibitive with increases in $T$. Therefore, instead of first performing regularized regression or logistic regression for the given parameters and then searching for the optimal combination of parameters, we propose conducting parameter tuning with cross-validation inside the algorithm. Specifically, we perform the cross-validation within each task at the initialization step. Hence, at the initialization step, we not only provide an initial estimator $\mathbf{W}^0$ but also find the best tuning parameter $\lambda_i$ for each task. The selected $\lambda_i$ are also used at the step of updating $\mathbf{W}$. Then, we set $\gamma$ as the average of $\lambda_i$'s: $\gamma=\frac{1}{T}\sum_{i=1}^T\lambda_{i}$, where $\lambda_i$ are the parameters determined at the initialization step. As a result, the tuning parameters have been selected inside the algorithm. Since we no longer need to run the algorithm multiple times over a grid space of penalty parameters, we can largely reduce the computational time. The effectiveness and efficiency of the proposed approach to selecting parameters have both been validated in both synthetic and real-world datasets.  
\begin{table}
	\begin{center}
		\scalebox{1}{
			\begin{tabular}{l}
			\toprule[2pt]
			\textbf{Algorithm 1} STCMTL \\
			\midrule[1pt]
			\textbf{Input:} Training datasets $\{\mathbf{X}_i,\mathbf{y}_i\}_{i=1}^T$ and number of clusters $K$ \\ 
		
			\textbf{Output:} $\mathbf{U}$ and $\mathbf{V}$ \\
			 
			\hline
			1: \textbf{Initialization}: $\mathbf{W}^0$; \\
			2: \textbf{Repeat} \\
			3: \quad Update the membership matrix $\mathbf{V}$ by using SOUP;\\
			4: \quad Update the cluster coefficient matrix $\mathbf{U}$ by minimizing 
			Eq. (\ref{U_update});\\
			5: \quad Update the coefficient matrix $\mathbf{W}$ by solving Eq. (\ref{single});\\
			6: \textbf{Until} the objective function of Eq. (\ref{main}) converges.\\
			\bottomrule[1pt]
		\end{tabular}
	}
	
	\end{center}
\end{table}

\section{Experiments}
\label{exp}
This section aims to assess the performance of the proposed STCMTL approach on both synthetic and real-world datasets. We compare STCMTL with the following baseline methods:  

\textbf{LASSO} [\cite{Lasso}]: The single-task learning method learns a sparse prediction model by applying LASSO for each task separately. 

\textbf{CMTL} [\cite{cluster2}]: This MTL method learns a disjoint cluster structure among the tasks and does not conduct feature selection. 

\textbf{Trace} [\cite{trace1}]: This MTL method learns a low-rank structure among the tasks by penalizing both the Frobenius norm and trace norm of the coefficient matrix. 

\textbf{FLARCC} [\cite{Fused}]: This MTL method achieves feature selection and coefficient clustering across tasks by using LASSO and fused LASSO penalties, respectively. 

\textbf{VSTG-MTL} [\cite{VSTG}]: This MTL method is the first method that simultaneously performs feature selection and learns an overlapping cluster structure among tasks using a low-rank approach.

These methods differ in their ability to perform feature selection, task clustering, and prediction, as shown in Table \ref{methods}. LASSO is implemented using the R package "glmnet", and CMTL and Trace using the R package "RMTL". The implementation of VSTG-MTL was released at https://github.com/ 
JunYongJeong/VSTG-MTL. STCMTL is implemented in a R package and is available at https://github.com/ RUCyuzhao/STCMTL. Suggested by [\cite{VSTG}], for VSTG-MTL, the penalty parameter $\mu$ of $k$-support norm is set equal to the penalty parameter $\gamma_1$ of $l_1$ norm, which reduces the computational burden. The hyper-parameters of all methods are selected via five-fold cross validation over a range of $\{2^{-15},\cdots,2^{3}\}$. The number of clusters for STCMTL, CMTL, and VSTG-MTLare selected from the search grid $\{2,3,...,9\}$.  

\begin{table}[]
	\centering
	\caption{STCMTL and baseline methods}
	\label{methods}
	\scalebox{0.85}{
		\begin{tabular}{lccc}
			\toprule[2pt]
			& Feature selection      & Task clustering       & Prediction   \\
			\hline
			STCMTL        &  $\surd$ & $\surd$ &$\surd$  \\
			LASSO           &  $\surd$ &  &$\surd$\\
			CMTL             &   & $\surd$ &$\surd$ \\
			Trace             &    &  & $\surd$ \\
			FLARCC   &$\surd$ & $\surd$ & $\surd$ \\
			VSTG -MTL& $\surd$ & $\surd$ & $\surd$ \\ 		
			\hline
	\end{tabular}}
\end{table}

We evaluate the prediction performance by using the root mean squared error (\textbf{RMSE}) for the regression problem and the error rate (\textbf{ER}) for the classification problem. For the synthetic dataset, we also evaluate the estimation and feature selection performance. Specifically, the estimation is quantified by using the root mean estimation error (\textbf{REE}), which is defined as $||\mathbf{W}-\mathbf{\hat{W}}||_F/\sqrt{T}$, whereas feature selection is quantified by using Matthew’s Correlation Coefficient (
\textbf{MCC}), which is defined as: 
$$ \mathrm{MCC}=\frac{\mathrm{TP} \times \mathrm{TN}-\mathrm{FP} \times \mathrm{FN}}{\sqrt{(\mathrm{TP}+\mathrm{FN})(\mathrm{TP}+\mathrm{FP})(\mathrm{TN}+\mathrm{FP})(\mathrm{TN}+\mathrm{FN})}}.$$  

\subsection{Synthetic Datasets}
We generate 60 tasks ($T=60$). For the $i$th task, 100 training observations ($n_i=100$) and 100 testing observations are generated from $\mathbf{X}_i \sim \mathcal{N}(0,\mathbf{I}_d)$ and $\mathbf{y}_i = \mathbf{X}_i \mathbf{w}_i+\mathcal{N}(0,0.5\mathbf{I}_{n_i})$. To simulate the high-dimensional scenario ($D>n_i$), which is more challenging than the low-dimensional one, we set $D=200$ and $600$. The 60 tasks are divided into five overlapping clusters ($K=5$), that is, the true coefficient matrix $\mathbf{W}=\mathbf{U}\mathbf{V}$, where $\mathbf{U} \in \mathcal{R}^{D \times K}$ and $\mathbf{V} \in \mathcal{R}^{K \times T}$. For each $\mathbf{W}$ we repeat the data generation process for 10 times and the average performances on the test sets are reported 

In the $k$th column of the cluster coefficient matrix $\mathbf{U}$, only the $[5(k-1)+1]$th to the $[5(k+1)]$th components are non-zero, which are generated from two uniform distributions, $\mathcal{U}[-0.5,-0.1]$ and $\mathcal{U}[0.1,0.5]$, randomly and independently; the other components are zeros. The 60 tasks consist of 50 pure tasks, equally assigned to $K$ clusters, and 10 mixed tasks. Without a loss of generality, we set the first 50 tasks as pure tasks and the last 10 as mixed tasks. As a result, for $i=1,\ldots,50$, the membership vectors $\mathbf{v}_i$ only have one component equal to 1, while the others are all equal to zero. For the last 10 mixed tasks, we consider two types of mixing patterns: (a) sparse, where the membership vectors $\mathbf{v}_i$ $(i=51,\ldots,60)$ have positive values for only two components, which are taken randomly from $K$ components; and (b) dense, where the membership vectors $\mathbf{v}_i$ $(i=51,\ldots,60)$ have positive values on all $K$ components. All membership vectors are normalized such that they sum to 1. 

\begin{figure}[htbp] 
	\centering
	\subfigure[Sparse]{\includegraphics[width=1.5in]{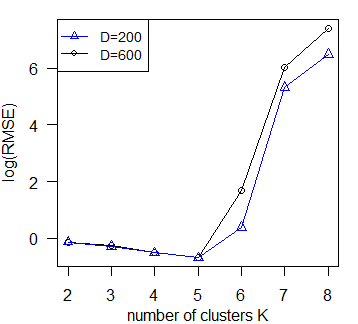}}
	\subfigure[Dense]{\includegraphics[width=1.5in]{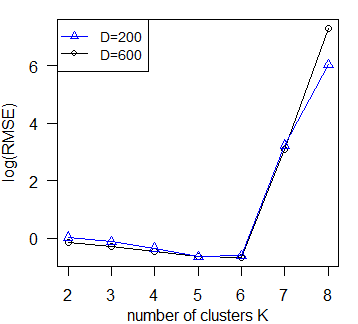}} 
	\caption{ Results for the synthetic datasets: logarithm of RMSE as a function of the number of clusters $K$.}
	\label{clusternumber}
\end{figure}

\begin{table*}[]
	\caption{Results for the synthetic datasets: RMSE, REE, and MCC based on 10 repetitions. Each cell shows the mean $\pm$ s.d. }
	\label{synthetic}
	\scalebox{0.75}{
		\begin{tabular}{lllllllll}
			\toprule[2pt]
			&&                                & LASSO         & CMTL        & Trace       &    FLARCC     & VSTG-MTL        & STCMTL          \\
			\hline
			\multirow{6}{*}{$D$=200} & \multirow{3}{*}{Sparse}     & RMSE & 0.659$\pm$0.015 & 0.651$\pm$0.007 & 0.911$\pm$0.001 & 0.591$\pm$0.011 & 0.543$\pm$0.005 & \bf{0.510$\pm$0.005} \\
			&                             & REE  & 0.030$\pm$0.001 & 0.029$\pm$0.006 & 0.054$\pm$0.001 & 0.019$\pm$0.001 & {0.011$\pm$0.001} & \bf{0.007$\pm$0.001} \\
			&                             & MCC  & 0.481$\pm$0.021 & 0.109$\pm$0.002 & 0.071$\pm$0.003 & 0.583$\pm$0.023 & 0.705$\pm$0.044 &\bf{0.779$\pm$0.018} \\
			& \multirow{3}{*}{Dense} & RMSE & 0.665$\pm$0.016 & 0.650$\pm$0.008 & 0.897$\pm$0.010 & 0.590$\pm$0.011 & \bf{0.515$\pm$0.007} & \bf{0.515$\pm$0.008} \\
			&                             & REE  & 0.031$\pm$0.001 & 0.029$\pm$0.006 & 0.053$\pm$0.004 & 0.019$\pm$0.001 & \bf{0.009$\pm$0.001} & \bf{0.009$\pm$0.001} \\
			&                             & MCC  & 0.470$\pm$0.018 & 0.111$\pm$0.003 & 0.071$\pm$0.002 & 0.613$\pm$0.080 & 0.722$\pm$0.025 & \bf{0.783$\pm$0.024} \\
			\hline
			\multirow{6}{*}{$D$=600} & \multirow{3}{*}{Sparse}     & RMSE & 0.743$\pm$0.015 & 0.829$\pm$0.011 & 1.038$\pm$0.013 & 0.635$\pm$0.019 & 0.525$\pm$0.010 & \bf{0.516$\pm$0.005} \\
			&                             & REE  & 0.022$\pm$0.000 & 0.027$\pm$0.000 & 0.036$\pm$0.000 & 0.016$\pm$0.000 & \bf{0.006$\pm$0.001} & 0.010$\pm$0.002 \\
			&                             & MCC  & 0.400$\pm$0.010 & 0.083$\pm$0.002 & 0.070$\pm$0.004 & 0.660$\pm$0.011 & 0.477$\pm$0.070 & \bf{0.715$\pm$0.157} \\
			& \multirow{3}{*}{Dense} & RMSE & 0.734$\pm$0.015 & 0.824$\pm$0.011 & 1.017$\pm$0.015 & 0.624$\pm$0.036 & \bf{0.525$\pm$0.010} & {0.526$\pm$0.006} \\
			&                             & REE  & 0.022$\pm$0.000 & 0.026$\pm$0.000 & 0.036$\pm$0.000 & 0.018$\pm$0.000 & {0.010$\pm$0.001} & \bf{0.005$\pm$0.001} \\
			&                             & MCC  & 0.386$\pm$0.013 & 0.088$\pm$0.002 & 0.073$\pm$0.006 & 0.643$\pm$0.059 & 0.542$\pm$0.084 & \bf{0.717$\pm$0.159}\\
			\hline
	\end{tabular}}
\end{table*}

With the proposed STCMTL approach, we first examine the effect of the number of clusters $K$ on RMSE. Fig. \ref{clusternumber} presents the logarithm of RMSE as a function of $K$ for a random replicate, where other hyper-parameters are selected by cross-validation inside the algorithm. RMSE reaches the minimum at the true number of clusters, which is five for the synthetic datasets. We also examine a few other replicates and observe similar patterns. 

\begin{figure}[]
	\centering
	\subfigure[True]{\includegraphics[width=0.3 \hsize]{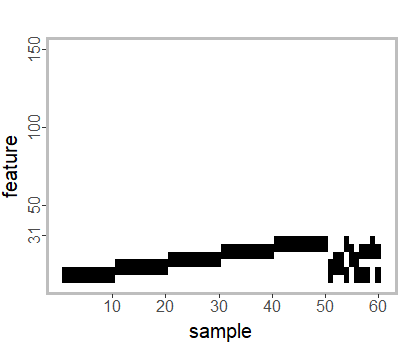}}
	\subfigure[STCMTL]{\includegraphics[width=0.3 \hsize]{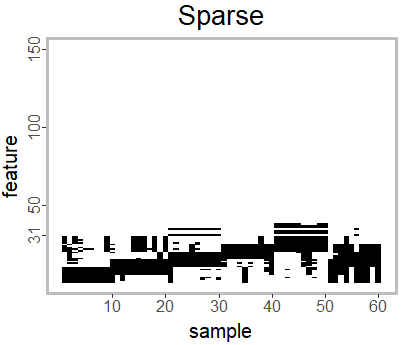}}
	\subfigure[VSTG-MTL]{\includegraphics[width=0.3 \hsize]{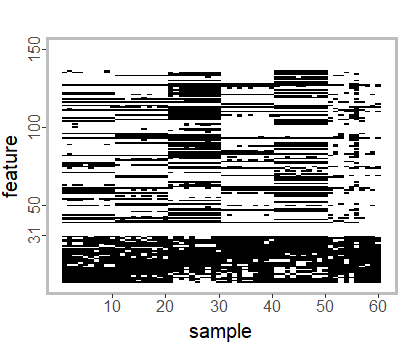}}\\
	\subfigure[True]{\includegraphics[width=0.3 \hsize]{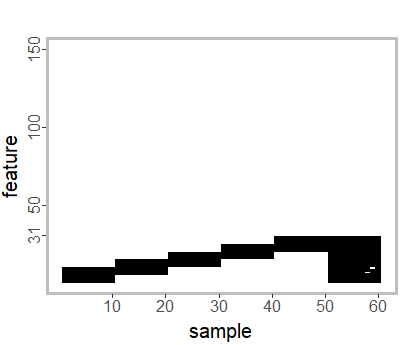}}
	\subfigure[STCMTL]{\includegraphics[width=0.3 \hsize]{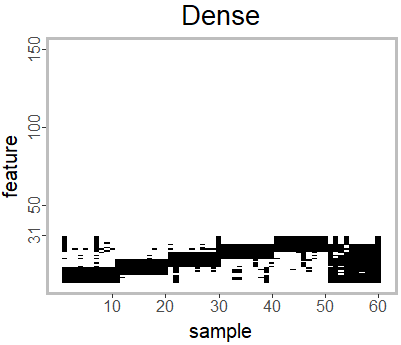}}
	\subfigure[VSTG-MTL]{\includegraphics[width=0.3 \hsize]{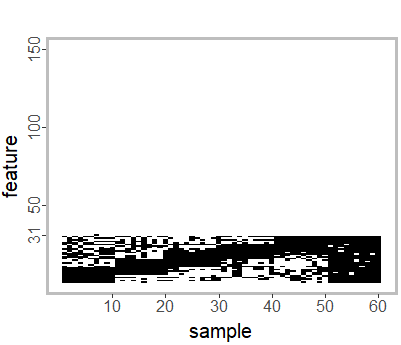}} 
	\caption{Results for the synthetic datasets: true and estimated coefficient matrices by STCMTL and VSTG-MTL. The dark- and white-colored entries indicate the nonzero and zero values, respectively.}
	\label{variable_selection}
\end{figure}
We then compute the summary statistics based on 10 repetitions. Table \ref{synthetic} presents the simulation results for the synthetic datasets. The values after $\pm$ are the standard deviations of the corresponding metrics values. In general, STCMTL outperforms the five baseline methods in terms of the three evaluation metrics in the majority of scenarios. STCMTL and VSTG-MTL perform significantly better than other methods. This is expected, as LASSO does not consider the task heterogeneity, CMTL and Trace do not conduct feature selection, and FLARCC clusters tasks based on each coefficient separately. For prediction, STCMTL always have the favorable performance, especially in Sparse condition. In terms of estimation, STCMTL has the best or nearly the best performance. Regarding the accuracy of feature selection, STCMTL performs significantly better than the baseline methods. VSTG-MTL provide a passable accuracy when $D=200$ but it clearly suffers when $D$ becomes large. The selection of correct features is crucial in some areas, such as biomedicine, which values the interpretability of results. VSTG-MTL, while it provides comparable prediction and estimation results, performs poorly in feature selection. Fig. \ref{variable_selection} shows the true coefficient matrix and estimated coefficient matrix by using STCMTL and VSTG-MTL with $D=600$, where the dark- and white-colored entries indicate non-zero and zero values, respectively \footnote{For demonstration purpose, we reorder the features and put the features selected by STCMTL or VSTG-MTL on the top of list.}. It can be seen that STCMTL can almost identify the true cluster structure among the tasks and remove most of the irrelevant features. However, VSTG-MTL suffers from selecting more false features, which inevitably affects the identification accuracy of the cluster structure.  

As VSTG-MTL is the method that is most relevant to the proposed STCMTL approach, we also compare the computational efficiencies of the two methods under all simulation settings. Table \ref{synthetic_time} shows the time spent in updating $\mathbf{U}$ and $\mathbf{V}$ with the fixed hyper-parameters, and completing the entire training process (including performing five-fold cross-validation), respectively, and their standard deviations for two approaches over 10 repetitions. Clearly, the computational efficiency of STCMTL is significantly higher than that of VSTG-MTL. The reasons for this are summarized as follows:  
\begin{itemize}
	\item STCMTL only contains a $\ell_1$ norm penalty term on $\mathbf{U}$, and thus, uses the coordinate descent algorithm, a highly efficient algorithm, to update $\mathbf{U}$. Meanwhile VSTG-MTL adopts the ADMM algorithm, a more complex algorithm, to address two penalty terms on $\mathbf{U}$. In particular, the time that VSTG-MTL spends on the updating process $\mathbf{U}$ grows significantly when the number of features $D$ increases from $200$ to $600$. 
	\item As described in Section \ref{alg}, STCMTL conducts the tuning of penalty parameters with cross-validation only at the initialization step, avoiding running the entire algorithm multiple times on the grid space of the penalty parameters. Hence, considerable time is saved.  
\end{itemize}

\begin{table*}[]
	\centering
	\caption{Results for the synthetic datasets: time consumption (in seconds) of STCMTL and VSTG-MTL based on 10 repetitions. Each cell shows the mean $\pm$ s.d.}
	\label{synthetic_time}
	\scalebox{0.8}{
		\begin{tabular}{lllllll}
			\toprule[2pt]
			& &                                  & Updating $\bf{U}$      & Updating $\bf{V}$      &  Total     \\
			\hline
			\multirow{4}{*}{$D$=200} & \multirow{2}{*}{Sparse}     & VSTG-MTL &13.0$\pm$2.3 & \bf{0.8$\pm$0.1} 
			&4448.7 $\pm$ 308.2 \\
			&                             & STCMTL      
			& \bf{5.1$\pm$0.8} &0.9$\pm$0.2 
			& \bf{56.9$\pm$7.2}
			\\
			& \multirow{2}{*}{Dense} & VSTG-MTL 
			&12.1$\pm$1.9 & \bf{0.8$\pm$0.1 }
			&	4532.9$\pm$191.6 \\
			&                             & STCMTL 
			&\bf{5.2$\pm$ 0.6} &0.9 $\pm$ 0.1
			&\bf{44.1}$\pm$\bf{3.6} 
			\\
			\hline
			\multirow{4}{*}{$D$=600} & \multirow{2}{*}{Sparse}     & VSTG-MTL 
			&141.2 $\pm$ 16.8 &\bf{1.1 $\pm$ 0.2}
			&39020.2 $\pm$ 1522.6
			\\
			&                             & STCMTL
			&\bf{48.3$\pm$3.8} & 2.1 $\pm$0.1
			&\bf{190.2 $\pm$ 20.8}
			\\
			& \multirow{2}{*}{Dense}& VSTG-MTL
			&128.6 $\pm$ 11.5 &\bf{1.0 $\pm$ 0.1 }
			&39169.5 $\pm$ 1921.5 \\
			&                             & STCMTL
			&\bf{45.3$\pm$ 3.9} &2.9 $\pm$0.2
			&\bf{191.4$\pm$20.2 }
			\\
			\hline
	\end{tabular}}
\end{table*}

In summary, from the synthetic datasets, we can see that STCMTL achieves the most desirable performance in terms of prediction, estimation, and feature selection compared to the five baseline methods. While VSTG-MTL is competitive in some scenarios, it suffers severely from poor feature selection and a prohibitive computational cost compared with STCMTL.

\subsection{Real-World Datasets}
We evaluate the performance of STCMTL on the following four real-world benchmark datasets. \textbf{Isolet data}\footnote{www.zjucadcg.cn/dengcai/Data/data.html} and \textbf{School data}\footnote{http://ttic.uchicago.edu/~argyriou/code/index.html} are two regression tasks datasets, which are widely used in previous works [\cite{isolet2,VSTG,robust}]. Training and test sets are obtained by splitting the dataset $75\%$-$25\%$. \textbf{MNIST data}\footnote{http://yann.lecun.com/exdb/mnist/} and \textbf{USPS data}\footnote{http://www.cad.zju.edu.cn/home/dengcai/Data/MLData.html} are both handwritten 10 digits datasets. We treat these 10-class classification as a multi-task learning problem where each task is the classification of one digit from all the other digits, thus yielding 10 tasks. Following the procedure in [\cite{overlap2,cluster2,VSTG}], each image is preprocessed with PCA, and the dimensionality is reduced to 64 and 87, respectively. For each task, the training (test) set is generated by randomly selecting 100 train (test) observations of the corresponding digit and 100 train (test) observations of other digits.

\begin{figure*}[htbp]
	\centering
	\subfigure[Isolet]{\includegraphics[width=0.24 \hsize]{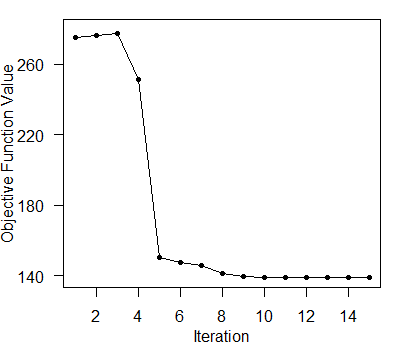}}
	\subfigure[School]{\includegraphics[width=0.24 \hsize]{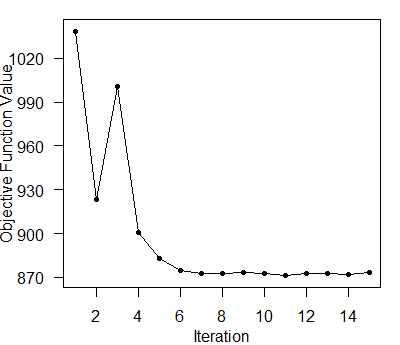}}
	\subfigure[MNIST]{\includegraphics[width=0.24 \hsize]{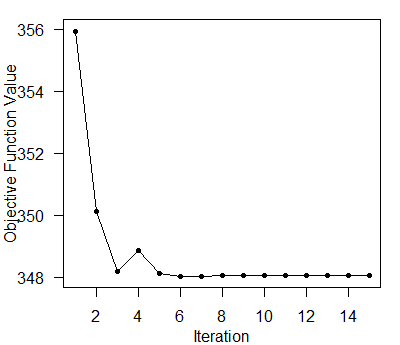}}
	\subfigure[USPS]{\includegraphics[width=0.24 \hsize]{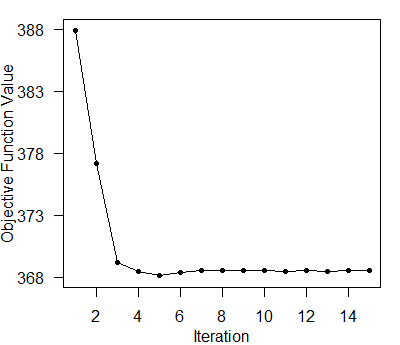}}
	\caption{Objective function value versus iteration on four real-world datasets.}
	\label{converge}
\end{figure*}

We apply STCMTL to the four real datasets. Fig. \ref{converge} shows the objective function value by iteration, and we can see that the objective function converges to the optimal value within 20 iterations in most repetitions. 

Table \ref{real} summarizes the prediction performance on the test sets of the five methods in the four real-world datasets over 10 repetitions. In the linear regression problem, STCMTL and VSTG-MTL stand out. The two methods' results are comparable, with no statistically significant differences. However, in the classification problem, the STCMTL outperforms all the other baseline methods. In addition, once again, the computational cost of VSTG-MTL is prohibitive compared to STCMTL (Table \ref{real_time}). 

\begin{table*}[]
	\caption{Results for the real-world datasets: RMSE and ER based on 10 repetitions. Each cell shows the mean $\pm$ s.d. }
	\label{real}
	\scalebox{0.8}{
		\begin{tabular}{lllllllll}
			\toprule[2pt]
			Dataset&                                & LASSO         & CMTL        & Trace       &  FLARCC       & VSTG-MTL        & STCMTL          \\
			\hline
			Isolet    & RMSE & 0.241$\pm$0.005 & 0.273$\pm$0.003 &0.237$\pm$0.003 & 0.262$\pm$0.006 & {0.193$\pm$0.018} & \bf{0.191$\pm$0.003}\\
			School &   & 10.638$\pm$0.031 & 10.564$\pm$0.056 & 12.210$\pm$0.053 & 10.771$\pm$0.059 & \bf{10.020$\pm$0.189} &{10.075$\pm$0.107} \\
			\hline
			MNIST & ER  & 0.202$\pm$0.013 & 0.187$\pm$0.007 & 0.165$\pm$0.007 & 0.190$\pm$0.009 & 0.108$\pm$0.012 & \bf{0.088$\pm$0.008} \\
			
			USPS&   & 0.196$\pm$0.011 & 0.199$\pm$0.010 & 0.170$\pm$0.009 & 0.184$\pm$0.012 & 0.101$\pm$0.009 & \bf{0.091$\pm$0.008} \\
			
			\hline
	\end{tabular}}
\end{table*}

\begin{table*}[]
	\centering
	\caption{Results for the real-world datasets: time consumption (in seconds) of STCMTL and VSTG-MTL based on 10 repetitions. Each cell shows the mean $\pm$ s.d. }
	\label{real_time}
	\scalebox{0.8}{
		\begin{tabular}{lll|ll|ll|ll}
			\toprule[2pt]
			& \multicolumn{2}{c}{Isolet} & \multicolumn{2}{c}{School} & \multicolumn{2}{c}{MNIST}    & \multicolumn{2}{c}{USPS}       \\
			& VSTG-MTL          & STCMTL        & VSTG-MTL        & STCMTL        & VSTG-MTL          & STCMTL           & VSTG-MTL            & STCMTL          \\
			\hline
			Total & 50029.8$\pm$3657.7   &\bf{525.3$\pm$35.9} & 934.1$\pm$95.6 & \bf{20.8$\pm$2.6}   & 2514.4$\pm$99.4 & \bf{209.3$\pm$30.2}    & 3077.8$\pm$154.3 & \bf{121.1$\pm$11.1}   \\
			\bottomrule 
	\end{tabular}}
\end{table*}

\section{Extension to Robust Task Clustering}
The proposed approach decouples task clustering and feature selection into estimating three tractable sub-problems: the membership matrix $\mathbf{V}$, cluster coefficients matrix $\mathbf{U}$, and coefficient matrix $\mathbf{W}$. The proposed approach employs the SOUP algorithm to identify the membership matrix but it is not the only available option, as numerous other matrix-factorization-based clustering methods have also been proposed. Hence, in practice, we could replace or combine SOUP with other matrix-factorization-based clustering methods to tackle more complex scenarios in task clustering. In this section, we consider task clustering with several outlier tasks and modify the proposed approach to a robust task clustering one.

We assume that there are some outlier tasks that cannot be represented by a linear combination of cluster coefficient vectors $\mathbf{v}_k$'s. To improve the robustness against these outlier tasks, in the step of updating $\mathbf{V}$, we first apply a $l_{12}$-norm-based robust nonnegative matrix factorization (NMF) algorithm to identify the outlier tasks. Then, we estimate $\mathbf{V}$ by using the SOUP algorithm. Specifically, we apply the robust NMF algorithm \cite{RobustNmf} on $\mathbf{W}$, yielding two rank-$K$ matrices, $\mathbf{C}$, and $\mathbf{\Theta}$, such that $\mathbf{W}\approx \mathbf{C}\mathbf{\Theta}$. Then, we define a distance $D_i=\|\mathbf{W}_i-\mathbf{C}\mathbf{\Theta}_i\|_2$ for each task, which measures the discrepancy between the true and estimated coefficients. If $\mathbf{W}$ is precise enough, the distances of the outlier tasks should be much larger than those of the normal tasks; thus, the outlier tasks can be identified. For simplicity, we employ the definition of outliers in bar plots. In words, a task is declared an outlier if it is above 4.5 times the interquartile range (IQR) over the upper quartile of the distance. Practically, we can use other advanced anomaly detection methods. Once the outlier tasks are identified, they are removed from the set of tasks. Then, we apply the SOUP algorithm on the remaining normal tasks to recover their membership weights. The overall procedure is summarized in Algorithm 2 \ref{Ar}, where $\mathbf{B}_\mathcal{S}$ refers to the sub-matrix of matrix $\mathbf{B}$ formed by columns in set $\mathcal{S}$. 

We use synthetic datasets to validate the effectiveness of the modified approach in terms of task clustering, feature selection, and robustness against outlier tasks. The data generation procedure is exactly the same as the sparse condition in section 4.1, except we add five outliers tasks, each of which has 10 nonzero coefficients generated from $\mathcal{U}[0.5,1]$. As a result, we have total $T=65$ tasks. We set $D = 100 \text{ and }200$. 
\begin{table}
	\begin{center}
		\begin{tabular}{l}
			\toprule[2pt]
			\textbf{Algorithm 2} Robust STCMTL\\
			\midrule[1pt]
			\textbf{Input:} Training datasets $\{\mathbf{X}_i,\mathbf{y}_i\}_{i=1}^T$, number of clusters $K$ and index of $T$ tasks $\mathcal{T}$;\\
			\textbf{Output:} matrix $\mathbf{U}$, $\mathbf{V}$ and the set of outlier tasks $\Gamma$ \\
			\hline
			1: \textbf{Initialization}: $\mathbf{W}^0$, $\Gamma=\emptyset$; \\
			2: \textbf{Repeat} \\
			3:\quad Apply the robust NMF algorithm on $\mathbf{W}_{\mathcal{T}\setminus \Gamma}$, yielding two rank-$K$
			matrices $\mathbf{C}$ and $\mathbf{\Theta}$, \\ such that $\mathbf{W}_{\mathcal{T}\setminus \Gamma}\approx \mathbf{C}\mathbf{\Theta}$; \\
			4:\quad For each task $i\in \mathcal{T}\setminus\Gamma$, compute its distance:  
			$D_i=\|\mathbf{W}_i-\mathbf{C}\mathbf{\Theta}_i\|_2$;\\
			5:\quad Update $\Gamma$ by: $\Gamma\leftarrow\Gamma\cup \{i:D_i\geq \text{the upper quartile of}\ \{D_i\}+4.5\times \text{IQR}\}$;  \\
			6:\quad Update the cluster coefficient matrix $\mathbf{V}_{\mathcal{T}\setminus \Gamma}$ by using SOUP; \\  
			7:\quad Update the cluster coefficient matrix $\mathbf{U}$ by minimizing Eq. (\ref{U_update});\\
			8:\quad Update the coefficient matrix $\mathbf{W}_{\mathcal{T}\setminus \Gamma}$ by solving Eq. (\ref{single});  \\
			9: \textbf{Until} the objective function of Eq. (\ref{main}) with tasks in $\Gamma$ removed 
			converges. \\
			\bottomrule[1pt]
		\end{tabular}
	\label{Ar}
	\end{center}
\end{table}

We compare the modified approach with the two baseline methods, LASSO and VSTG-MTL. Table \ref{robust} reports the simulation results for the three methods. It can be observed that the modified approach outperforms the two baseline methods in terms of prediction, estimation, and feature selection. 

\begin{table}[]
	\centering
	\caption{Results for the synthetic datasets with outlier tasks: RMSE, REE, and MCC based on 10 repetitions. Each cell shows the mean $\pm$ s.d.}
	\scalebox{0.85}{
	\begin{tabular}{lllll}
		\toprule[2pt]
		&      & LASSO      & VSTG-MTL        &  STCMTL (Robust)    \\
		\hline
		& RMSE & 0.836$\pm$0.014 & 0.884$\pm$0.046 & \bf{0.577$\pm$0.029} \\
		$D$=100 &REE   & 0.067$\pm$0.002 & 0.073$\pm$0.003 & \bf{0.029$\pm$0.006} \\
		& MCC  & 0.413$\pm$0.012 & 0.647$\pm$0.064 &  \bf{0.660$\pm$0.035} \\
		\hline
		& RMSE & 1.011$\pm$0.019 & 0.878$\pm$0.046 & \bf{0.685$\pm$0.038} \\
		$D$=200 &REE   & 0.125$\pm$0.001 & 0.073$\pm$0.004 &  \bf{0.033$\pm$0.004} \\
		& MCC  & 0.331$\pm$0.078 & 0.637$\pm$0.066 & \bf{0.731$\pm$0.017}\\ 
		\hline
	\end{tabular}
}
	
	\label{robust}
\end{table}

\section{Conclusion}
This study proposes a novel task clustering approach STCMTL, which can simultaneously learn an overlapping structure among tasks and perform feature selection. 
Although it is similar in spirit to the existing task clustering methods, it is also significantly more advanced. In the classification of pure and mixed tasks, it produces an identifiable and interpretable task cluster structure and selects relevant features. 

STCMTL factorizes the coefficient matrix into a product of the cluster coefficient matrix and membership matrix, and it imposes some constraints on the membership matrix values and sparsities on the cluster coefficient matrix. T4edhe resulting constrained optimization problem is challenging to solve. To circumvent this, a new alternating algorithm is developed, which repeats a three-step process: identifying the set of pure tasks and estimating the membership matrix; updating the cluster coefficient matrix; and updating coefficient matrix until convergence. The tuning of hyper-parameters with validation is conducted inside the algorithm, leading to substantial improvements in computational efficiency. The experimental results show that the proposed STCMTL approach outperforms the baseline methods using synthetic and real-world datasets. Moreover, STCMTL can be extended to other, more complex task clustering problems, such as robust task clustering. 

An important limitation of this study is its lack of theoretical investigation. This will be deferred to future research. We also defer additional possible extensions of the proposed approach to future research. 

\newpage
\bibliography{acml_bib}

\end{document}